\documentclass[preprint,12pt]{elsarticle}

\usepackage{amssymb}

\usepackage{hyperref}
\usepackage{mathrsfs}
\usepackage{booktabs}
\usepackage{multirow}
\usepackage{amsmath}
\usepackage{bbding}
\usepackage{float}
\usepackage{graphicx}
\usepackage[export]{adjustbox}

\journal{Knowledge-Based Systems}

\begin{document}

\begin{frontmatter}







\title{A semantically enhanced dual encoder for aspect sentiment triplet extraction \tnoteref{1}}
\tnotetext[1]{This work was supported by Jiangsu Provincial Social Science Foundation of China (Grant No. 21HQ043).}
\author[a]{Baoxing Jiang}
\ead{wflqjbx158@126.com}
\author[b]{Shehui Liang \corref{cor}}
\ead{liangshehui@njnu.edu.cn}
\author[a]{Peiyu Liu \corref{cor}}
\ead{liupy@sdnu.edu.cn}
\author[a]{Kaifang Dong}
\author[a]{Hongye Li}
\affiliation[a]{organization={School of Information Science and Engineering}, 
            addressline={Shandong Normal University}, 
            city={Jinan},
            postcode={235000}, 
            country={China}}          
\affiliation[b]{organization={International College for Chinese Studies}, 
	addressline={Nanjing Normal University}, 
	city={Nanjing},
	postcode={210097}, 
	country={China}}
\cortext[cor]{Corresponding author}

\begin{abstract}
Aspect sentiment triplet extraction (ASTE) is a crucial subtask of aspect-based sentiment analysis (ABSA) that aims to comprehensively identify sentiment triplets. Previous research has focused on enhancing ASTE through innovative table-filling strategies. However, these approaches often overlook the multi-perspective nature of language expressions, resulting in a loss of valuable interaction information between aspects and opinions. To address this limitation, we propose a framework that leverages both a basic encoder, primarily based on BERT, and a particular encoder comprising a Bi-LSTM network and graph convolutional network (GCN ). The basic encoder captures the surface-level semantics of linguistic expressions, while the particular encoder extracts deeper semantics, including syntactic and lexical information. By modeling the dependency tree of comments and considering the part-of-speech and positional information of words, we aim to capture semantics that are more relevant to the underlying intentions of the sentences. An interaction strategy combines the semantics learned by the two encoders, enabling the fusion of multiple perspectives and facilitating a more comprehensive understanding of aspect--opinion relationships. Experiments conducted on benchmark datasets demonstrate the state-of-the-art performance of our proposed framework.
\end{abstract}


%
%


\end{frontmatter}


\section{Introduction}
\begin{figure}[H]
	\centering
	\scalebox{1}{
		\includegraphics{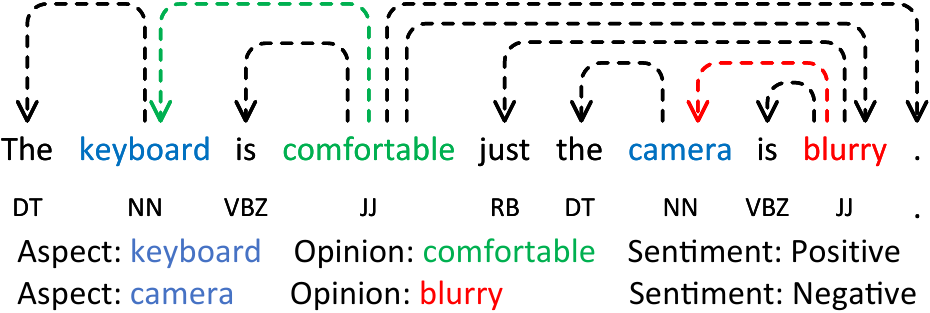}
	}
	\caption{An ASTE example with dependencies, where aspects, opinions, and sentiments are marked with different colors.}
	\label{example}
\end{figure}
Aspect sentiment triplet extraction (ASTE) \citep{peng-two-stage} is the most comprehensive subtask within aspect-based sentiment analysis (ABSA) \citep{semeval-2014-task4}, integrating the extraction, matching, and classification subtasks to simultaneously obtain aspects, opinions, and their corresponding sentiment attitudes.

ABSA was first accomplished through several independent subtasks, including aspect terms extraction  (ATE) \citep{ate1,ate2,ate3} and opinion terms extraction (OTE) \citep{ote1,ote2}, which focused solely on extracting aspect or opinion entities from sentences. Aspect-oriented opinion extraction (AOE) \citep{aoe1,aoe2,aoe3} aimed to match aspect--opinion pairs in comments, while aspect-level sentiment classification (ALSC) \citep{alsc1,alsc2,alsc3} determined affective attitudes based on given aspects in a sentence. However, the performance of these individual subtasks deteriorated when stacked.

\citet{peng-two-stage} defined the ASTE task as a sentiment triplet consisting of an aspect term, opinion term, and their corresponding sentiment, with the objective to extract triplets from sentences in the format [Aspect, Opinion, Sentiment]. Figure \ref{example} shows a comment about a laptop that addresses “keyboard" and “camera,"  so the triplets are [keyboard, comfortable, Positive] and [camera, blurry, Negative]. \citet{peng-two-stage}, who adopted a pipeline approach to the ASTE task, proposed a two-stage model. Aspects, opinions, and sentiments are extracted from sentences in the first stage, and combined in the second stage to generate appropriate triplets. However, the pipeline approach fragments the relation between the elements in a triplet, failing to capture their rich interactions.

A more advanced alternative, the joint model \citep{jet,span-aste,gts,emc-gcn,mao-joint,jing-joint,bdtf}, aims to capture interactions between subtasks in a unified manner, using either a table-filling or span-based strategy. In the table-filling approach, \citet{jet} pioneered the conversion of ASTE tasks into sequence tagging tasks using a BIOES position-aware tagging scheme. \citet{gts} abstracted sentences into a 2D table and employed specific symbols to label the cells in the grid for triplet extraction. To address issues such as relation inconsistency and boundary insensitivity, \citet{bdtf} proposed boundary-driven table-filling (BDTF). The table-filling approach allows for the learning of fine-grained semantics for each token in an end-to-end manner. However, it struggles to capture continuous features for multi-word entities, an area where span-based approaches excel. \citet{span-aste} proposed a span-based method to tackle the end-to-end model’s limitations in handling multi-word entities, where a sentence is spanned, and candidate aspect and opinion spans are generated by aspect terms extraction (ATE) and opinion terms extraction (OTE) modules. The matching module then determines the possible span pairs.

Whether applying a pipeline or end-to-end approach, the above work focuses on creating different extraction schemes to address ASTE, often neglecting a smartly engineered encoder to enhance the capture of high-order semantics. Upon observation, it becomes evident that high-order semantics can be both basic and specific. Previous research \citep{dependencyenhanced,posaware} has demonstrated that an effective encoder can consider specific semantics, including hidden features such as syntactic dependencies, part-of-speech tags, and position information. Higher-order information within a sentence can be represented by additional semantics between tokens. For example, in Figure  \ref{example}, the aspect–opinion pairs are lexically "noun–adjective" pairs, while there is an explicit syntactic dependency between an aspect and its corresponding opinion. On the other hand, the same word can also embody different linguistic meanings in different contextual expressions, as illustrated in Figure  \ref{example}. The adjective "blurry" is used negatively to describe "camera," whereas it may be used positively to describe another product. These rules hold true in a general sense.

To address the particular semantics mentioned above, a conventional Bi-LSTM \citep{lstm} encoder can effectively model implicit grammatical relationships. For basic semantics, BERT \citep{bert} encoders have shown success in encoding contextual information. However, fully capturing and refining higher-order semantics from multiple perspectives can further enhance the expressiveness of extraction schemes.

To harness the advantages of high-order semantics, we propose a semantically enhanced dual-encoder framework. We construct a BERT encoder to encode the basic semantics of a sentence, and employ a Bi-LSTM encoder augmented with 3-domain embeddings to extract particular semantics, which can differentiate part-of-speech and semantic differences in various contexts. For the obtained semantics, a GCN \citep{gcn} is employed to refine structural features based on the dependency tree. An encoding interaction module iteratively fuses the semantics of the text from these two perspectives.

Differing from current pipeline- and span-based methods, our approach extracts sentiment triplets based on boundary-driven table-filling (BDTF) \citep{bdtf}. Our model uses advanced coders to obtain inputs with richer contextual information and more comprehensive hidden semantic features, leading to improved performance on benchmark datasets. Experiments on publicly available datasets demonstrate our model’s state-of-the-art performance on the ASTE task.

The contributions of our paper are as follows.
\begin{enumerate}
	\item We propose a semantically enhanced  dual-encoder framework tailored for ASTE, comprising two distinct encoders. A basic encoder based on BERT captures the fundamental semantics of the sentence, and a particular encoder based on GCN and Bi-LSTM with 3-domain embeddings mines context-specific semantics from the dependency tree;
	\item We introduce a semantic encoding interaction strategy that iteratively combines the semantic information learned by the two encoders to comprehensively represent the sentence’s semantics, incorporating the basic context, specific review context, part-of-speech information, and dependency information;
	\item 3.	Our proposed framework outperforms baselines in terms of ASTE performance in experiments using two versions of the SemEval Challenges datasets\footnote{To facilitate related research, our code is publicly available at \url{https://github.com/BaoSir529/dual-encoder4aste}}.
\end{enumerate}

Overall, our contributions include the introduction of a semantic enhanced dual-encoder framework, a novel semantic encoding interaction strategy, and experimental validation showcasing the effectiveness of our approach on benchmark datasets.

\section{Proposed framework}

\begin{figure}
	\centering
	\scalebox{0.3}{
		\includegraphics{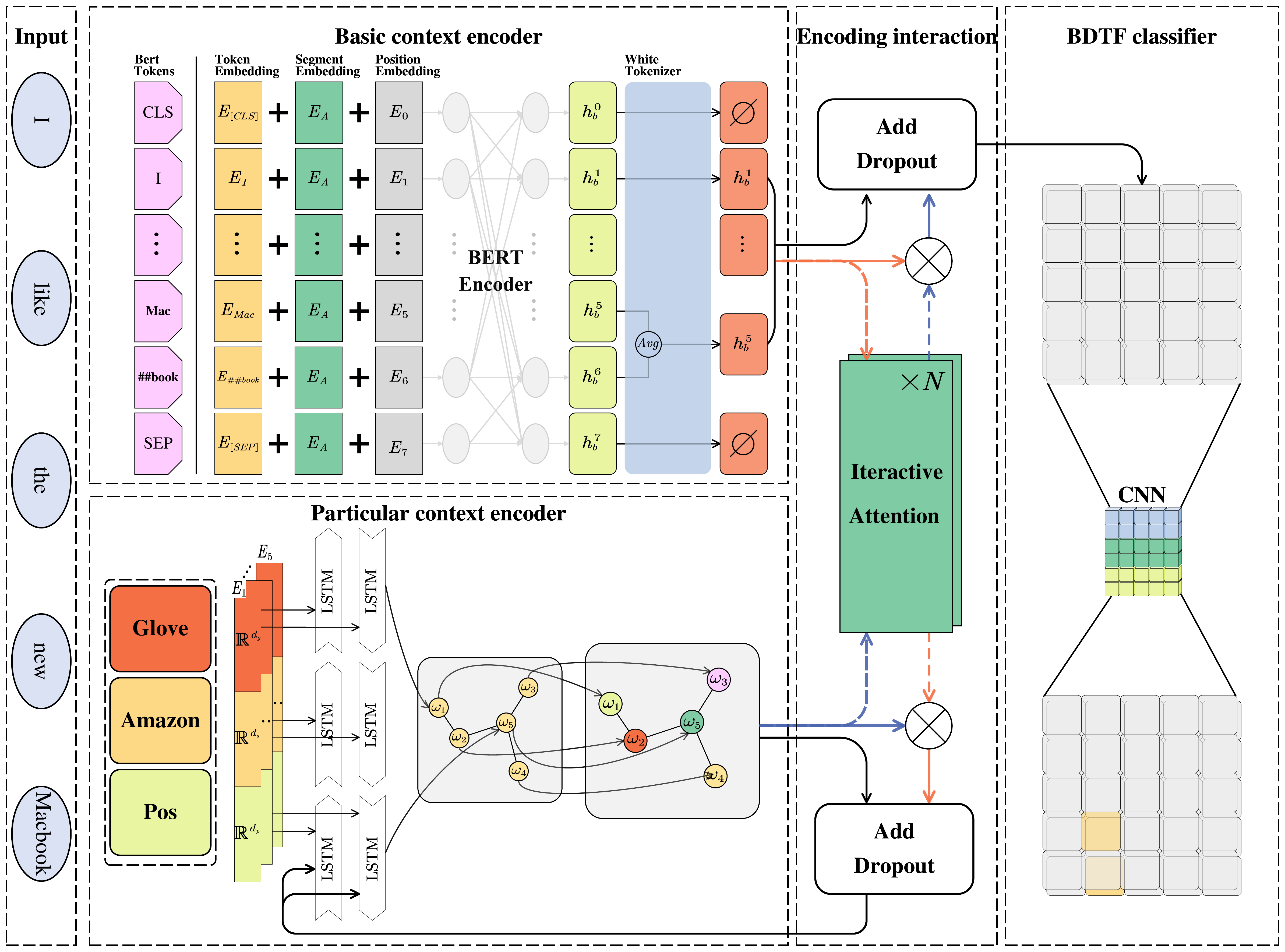}
	}
	\caption{The overview of our semantic enhanced dual-encoder.}
	\label{model}
\end{figure}

\subsection{Task definition}
Given a sentence $\mathcal{W}=\{\omega_1, \omega_2,\dots,\omega_n\}$ with $n$ words, the aspect sentiment triplet refers to the aspects $A$ described by the sentence, opinions $O$ expressed in it, and corresponding sentiment $S$ $\in$ \{Positive, Neutral, Negative\}. The goal of ASTE is to identify all the triples $T=\{(a_i,o_i,s)|(a_i,o_i,s) \in A\times O\times S\}$, where both $a_i\in A$ and $o_i \in O$ consist of one word or consecutive words in $\mathcal{W}$.

\subsection{Semantically enhanced  dual encoder}

Figure \ref{model} shows the four key components of our proposed framework: a basic context encoder, particular context encoder, encoding interaction module, and BDTF classifier.

We input a sentence to the basic context encoder, which uses BERT to encode the basic semantics. We employ 3-domain embeddings to perform word2vec  on the sentence and feed it into the particular context encoder to learn the semantics of the sentence within a given context. We simultaneously use a dependency tree to analyze the grammatical information, and apply graph convolution to capture the syntactic dependencies. After passing through both encoders, we obtain two types of semantic representations for the sentence. These are fed into the encoding interaction module, which employs a multi-headed attention mechanism with nonlinear activation to establish comprehensive associations between the two encodings, thus capturing the synergistic effects of the basic and particular semantics. The fused representation is used to generate a labeled grid, from which the BDTF classifier extracts sentiment triplets.

\subsection{Basic context encoder}

A BERT encoder obtains the basic contextual semantics of sentences. The input is a sentence representation with the form "[CLS] sentence [SEP]," where [CLS] and [SEP] are tokens added by BERT at the beginning and end, respectively, of the sentence.

Since BERT uses sub-word splitting, the obtained contextual features h may not align with the original sentence length in terms of dimensions. To address this issue, we designed a Whitetokenizer, which keeps track of each word being split into subwords during tokenization. We apply averaging pooling to perform alignment on the subwords, aggregating their contextual features to obtain the aligned contextual semantics $h_b$ for the sentence. The process can be described as

\begin{equation}\label{bert}
	h_b={\rm BERT}({\rm W_{tok}}([{\rm [CLS]},\omega_1, \dots,\omega_n,{\rm [SEP]}]))
\end{equation}
\begin{equation}
	h_b\left[ k \right] =\frac{1}{j-i}\sum_{n=i}^j{h_b[n]},\quad if\ [\omega_i; \omega_j] \in \omega _k.		
\end{equation}

It is worth noting that we remove the hidden state of unique markers [CLS] and [SEP], where W$_{\rm tok}$ denotes the designed whitetokenizer, $d_b \in \mathbb{R}^{n\times d_b}$ denotes the hidden state dimension of BERT, and $n$ is the sentence length.

\subsection{Particular context encoder}
Research \citep{dependencyenhanced,doubleembedding} has demonstrated that high-quality word embeddings are effective in capturing the semantic nuances of words within a sentence. These semantic distinctions are particularly evident in different expression contexts. In general domains, words exhibit varying semantics, and each word can have multiple meanings. However, in specific domains, the semantics of words tend to be biased toward  a particular aspect based on the context in which they are used.

To accurately capture the semantic tendencies of words in different domains, we employ two lookup tables to initialize the text embeddings. The general domains embedding table $E_g \in \mathbb{R}^{\left| v \right| \times d_g}$ represents the meanings of words in a broad context, while the specific domains embedding table $E_s \in \mathbb{R}^{\left| v \right| \times d_s}$ captures them in a context specific to the comments being analyzed. Here, $\left| v \right|$ denotes the size of the vocabulary in the lookup tables.

In addition to semantic information, part-of-speech (POS) plays a crucial role in distinguishing between words from a linguistic perspective. As illustrated in Figure \ref{example}, words with different POS tags exhibit specific collocations and distinctions. To incorporate the POS properties of words into our model, we use the SpaCy library\footnote{\url{https://spacy.io}} to obtain the POS tag for each word in a sentence. Since the full range of POS categories is extensive and overlapping, we simplify the POS tags into a set $\mathcal{P} = [p_{noun}, p_{verb}, p_{adj}, p_{adv}, p_{others}]$, whose elements represent nouns, verbs, adjectives, adverbs, and other categories, respectively. We introduce a learnable POS embedding matrix $E_p \in \mathbb{R}^{5 \times d_p}$ to encode the POS information of each word. Consequently, after passing through the 3-domain embedding layer, the final sentence representation $E \in \mathbb{R}^{n \times (d_g+d_s+d_p)}$ can be formulated as
\begin{equation}
	E=E_g \oplus E_s \oplus E_p,
\end{equation}
where, $\oplus$ denotes concatenation, and $n$ is the number of words in the sentence.

To capture the contextual features of sentences, we use Bi-LSTM as the particular encoder, which takes sentence embedding sequences $E$ as input and employs gating and attention mechanisms to obtain contextual features $X \in \mathbb{R}^{n \times d_l}$, where $d_l$ is the dimension of the hidden state. The process can be written as
\begin{equation}
	X = {\rm Bi\mbox{-}LSTM}(E).
\end{equation}

The feature matrix $X$ reflects the contextual features of the sentence. However, the grammatical information is also important. To incorporate this information, we employ a dependency tree implemented on a GCN, which learns the grammar of the comments and enhances the current representation by capturing the associations among words. In our implementation, the result $X$ from Bi-LSTM serves as the initial state. Multiple layers of graph convolution operations are performed on the corresponding dependency tree to obtain the grammar-enhanced contextual representation $h_p$. The process can be expressed as
\begin{equation}\label{gcnh}
	h^{(l)}_p = \sigma(\tilde{D}^{-\frac{1}{2}}\tilde{A}\tilde{D}^{-\frac{1}{2}}h^{(l-1)}_pW^{(l-1)}),
\end{equation}
where $h^{(l-1)}_p$ represents the features of each layer, and for the input layer, $h^0_p = X$. $W$ denotes the trainable parameters for each layer. $\tilde{A}$ is a $0\mbox{-}1$ adjacency matrix that records the connectivity between words, and $\tilde{D}$ is the degree matrix, which records the number of neighbors for each node. $\tilde{A}$ and $\tilde{D}$ are generated as
\begin{equation}
	\tilde{A}_{ij}=\left\{ \begin{array}{l}
		1,\ {\rm if}\ \omega _i\ {\rm connect\ with}\ \omega _j;\\
		0,\ {\rm otherwise;} \\
	\end{array} \right. 
\end{equation}
\begin{equation}
	\tilde{D}_{ii}=\sum\nolimits_{j}{\tilde{A}_{ij}}.
\end{equation}

Applying the above equations, we obtain the basic contextual features $h_b$ learned by the BERT-based encoder, and the particular contextual features $h_p$ learned by the Bi-LSTM encoder based on the 3-domain embedding. These are fed into the encoding interaction module to fuse the contextual features obtained by the two encoders.

\subsection{Encoding interaction module}
\label{representation}
The basic semantic $h_b$ reflects the hidden state of each word based on the contextual semantics. The particular semantic $h_p$ reflects the semantics of the sentence specific to the context of the comment domain, the lexical differences of each word and the syntactic differences of the sentence. The two semantics reflect the characteristics of the sentence from different perspectives. To integrate the two kinds of information with different emphases, a coding interaction module is designed to enhance the semantic features. 

First, the basic semantics $h_b$ and the particular semantics $h_p$ are fed into an interactive attention module to obtain respective attention scores $\alpha_b$ and $\alpha_p$ for each word,
\begin{equation}
	\alpha_b={\rm softmax}(h_bh_b^T)
\end{equation}
\begin{equation}
	\alpha_p={\rm softmax}(h_ph_p^T).
\end{equation}

Theoretically, $\alpha_b \in \mathbb{R}^{n \times n}$ records the semantic correlation between any words in a sentence, reflecting how close they are in terms of semantics, and similarly, $\alpha_p \in \mathbb{R}^{n \times n}$ carries much information about the lexical and grammatical relevance of the reflective words. To interact with the two types of semantics and enhance the expression, we let the attention fraction $\alpha_b(\alpha_p)$ act on the opposing semantic feature $h_p(h_b)$ to fuse the two semantics while reinforcing the original semantic feature by self-looping,
\begin{equation}
	h_b'={\rm Dropout}(\alpha_p h_b) + h_b
\end{equation}
\begin{equation}
	h_p'={\rm Dropout}(\alpha_b h_p) +h_p.
\end{equation}

The dropout layers are set to randomly mask some channels to resist gradient propagation errors and improve model robustness.

An attention mechanism enables the interaction of two sides of the semantics. However, a single layer of interaction does not sufficiently integrate the semantics. We introduce a multi-layer interaction mechanism to further strengthen the features,
\begin{equation}
	h^l_p={\rm Bi\mbox{-}LSTM}(g^{(k)}(h_p', \tilde{A}))
\end{equation}
\begin{equation}
	g^{(t)}(h_p', \tilde{A})=\sigma(\tilde{D}^{-\frac{1}{2}}\tilde{A}\tilde{D}^{-\frac{1}{2}}h'^{(t-1)}_pW^{(t-1)})
\end{equation}
\begin{equation}
	h^l_b=h_b',
\end{equation}

where $h^l_p$ and $h^l_b$ are the output of the $l$-th layer interaction, $g^k$ represents a $k$-layer GCN, $h'^{(t\mbox{-}1)}_p$ is the output of layer $t\mbox{-}1$  in the GCN, and $W^{(t\mbox{-}1)}$ is a trainable parameter of the $(t\mbox{-}1)$-layer GCN.

In the above interaction, $h^l_p$ can repeatedly extract valid information from the BERT base encoder and fuse itself through the interactive attention mechanism. Similarly, $h^l_b$ can attend to differences in particular semantics $h_p$ regarding lexical and syntactic properties while reinforcing self-features through interactive attention. Through iterative interactions of Eqs. \ref{bert}--\ref{gcnh}, we select the result $h^L_b$ of the $L\mbox{-}$th-layer interaction as the output of the coded interaction module, which is fed into a BDTF-based triplet extractor.

\subsection{Boundary-driven Table-filling layer}
\begin{figure}[H]
	\centering
	\scalebox{0.7}{
		\includegraphics{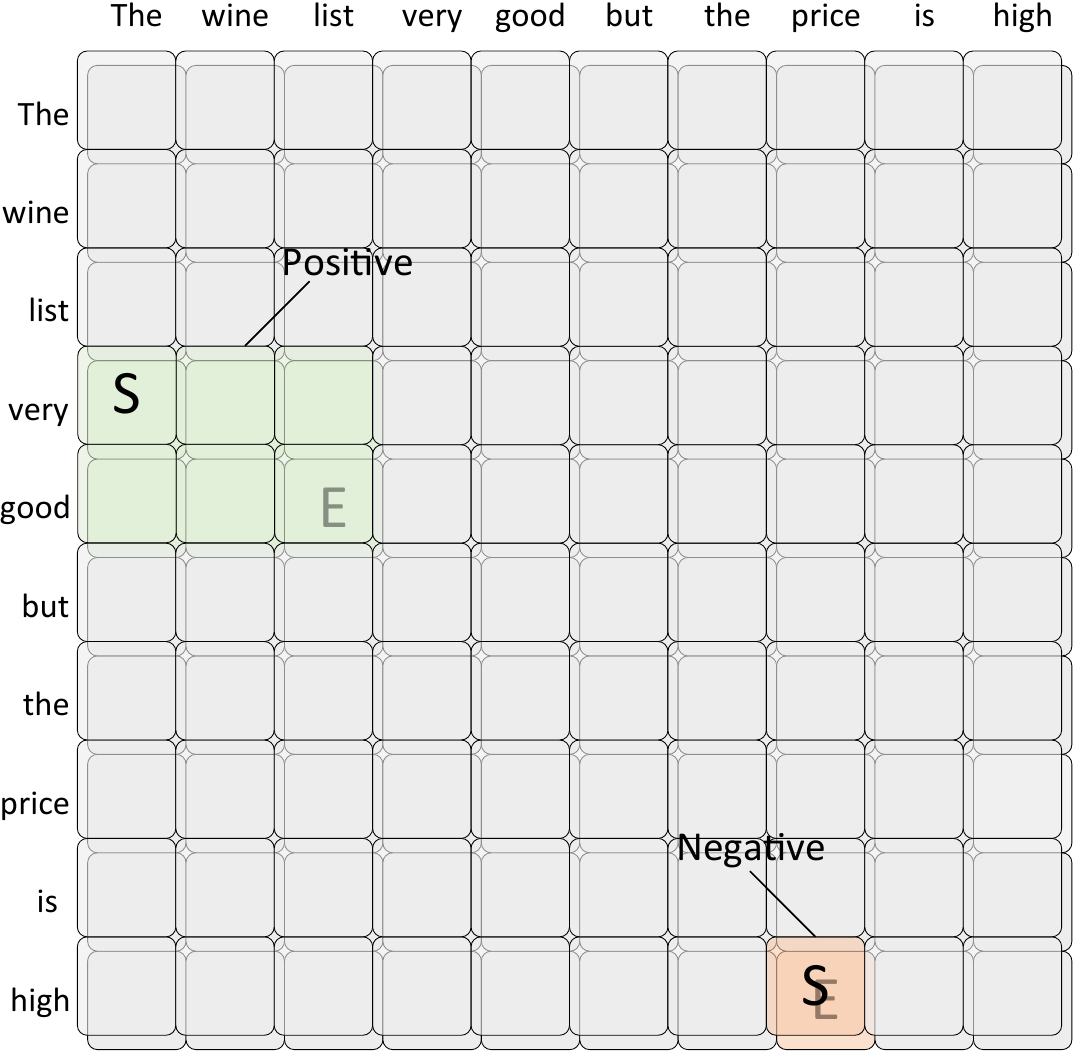}
 	}
	\caption{An example of BDTF-tagged sentence containing two triplets, (The wine list, very good, Positive) and (price, high, Negative) respectively.}
	\label{bdtfgrid}
\end{figure}

Boundary-driven table-filling (BDTF\footnote{For more details, please refer to \citet{bdtf}}) is a high-performance strategy for ASTE tasks, where, as shown in Figure \ref{bdtfgrid}, the aspect--opinion pair is represented as a two-channel bounded region, localized by the region's top-left start position $S[a_s,o_s]$ and bottom-right end position $E[a_e,o_e]$. Based on this, ASTE is converted into a multichannel region extraction and classification task.

For the high-dimensional semantics, followed by \citet{bdtf}, we construct the relationship-level representation $r_{ij}$ between random words 
\begin{equation}
	r_{ij}=\sigma({\rm Linear} ({\rm Pooling}(h_b^L[i],\dots,h_b^L[j]))),
\end{equation}
where $\sigma$ is an activation function, followed by \citet{gelu}, we use $gelu$.

\begin{table}
	\centering
	\scalebox{0.75}{
		\begin{tabular}{@{}lllllllllllllllll@{}}
			\toprule
			\multirow{2}{*}{Dataset} & \multicolumn{4}{c}{L\small AP\normalsize14}   & \multicolumn{4}{c}{R\small EST\normalsize14}  & \multicolumn{4}{c}{R\small EST\normalsize15} & \multicolumn{4}{c}{R\small EST\normalsize16}  \\ \cmidrule(l){2-17} 
			& $\#S$ & $\#A$  & $\#O$  & $\#T$  & $\#S$ & $\#A$  & $\#O$  & $\#T$  & $\#S$  & $\#A$ & $\#O$ & $\#T$  & $\#S$ & $\#A$  & $\#O$  & $\#T$  \\ \midrule
			\textbf{ASTE-Data-V1}          &        &      &      &      &        &      &      &      &         &     &     &      &        &      &      &      \\
			Train                    & 920    & 1283 & 1265 & 1265 & 1300   & 2079 & 2145 & 2145 & 593     & 834 & 923 & 923  & 842    & 1183 & 1289 & 1289 \\
			Dev                      & 228    & 317  & 337  & 337  & 323    & 530  & 524  & 524  & 148     & 225 & 238 & 238  & 210    & 291  & 316  & 316  \\
			Test                     & 339    & 475  & 490  & 490  & 496    & 849  & 862  & 862  & 318     & 426 & 455 & 455  & 320    & 444  & 465  & 465  \\ \midrule
			\textbf{ASTE-Data-V2}             &        &      &      &      &        &      &      &      &         &     &     &      &        &      &      &      \\
			Train                    & 906    & 1280 & 1254 & 1460 & 1266   & 2051 & 2061 & 2338 & 605     & 862 & 935 & 1013 & 857    & 1198 & 1300 & 1394 \\
			Dev                      & 219    & 295  & 302  & 346  & 310    & 500  & 497  & 577  & 148     & 213 & 236 & 249  & 210    & 296  & 319  & 339  \\
			Test                     & 328    & 463  & 466  & 543  & 492    & 848  & 844  & 994  & 322     & 432 & 460 & 485  & 326    & 452  & 474  & 514  \\ \bottomrule
		\end{tabular}
	}
	\caption{Statistics of the ASTE-Data-V1 and ASTE-Data-V2 datasets. $\#S$, $\#A$, $\#O$, and $\#T$ indicate the number of sentences, aspects, opinions, and triplets, respectively.}
	\label{dataset_statistic}
\end{table}

For a sentence of length $n$, the relation-level representation $r_{ij} \in \mathbb{R}^d$ between any two words forms a 3D relation matrix $R \in \mathbb{R}^{n\times n\times d}$, which is fed into $L$-layers ResNet-style CNN \citep{resnet} encoding layers to extract high-dimensional relations,

\begin{equation}
	R^{(l)}=\sigma({\rm Conv}(R^{(l\mbox{-}1)}))+R^{(l\mbox{-}1)},
\end{equation}
where $\sigma$ is an activation function (e.g., $ReLU$), and $Conv$ represents Convolution. We choose the output $R^{(L)}$ of the last CNN layer as the extraction result.

After a boundary detection classification layer, a pool of potential region candidates $P$ is obtained, which records the locations of predicted aspect–opinion pairs $[S(a_s,o_s ),E(a_e,o_e )]$ and their affective attitudes $\mathcal{S}$,

\begin{equation}
	P = {[(S^1,E^1, \mathcal{S}^1),\dots, (S^v,E^v, \mathcal{S}^v)]}
\end{equation}

\begin{align}
	S^i[a_s,o_s] 	&= {\rm top}\mbox{-} {\rm k}({\rm sigmoid}({\rm Linear}(R^{(L)})))\\
	E^i[a_e,o_e] 	&= {\rm top}\mbox{-} {\rm k}({\rm sigmoid}({\rm Linear}(R^{(L)})))
\end{align}

\begin{equation}
	\mathcal{S}^i={\rm softmax}({\rm Linear}([r_{a_so_s};r_{a_eo_e};c]))
\end{equation}

\begin{equation}
	c = {\rm Pooling}
	\left(	
	\left[ 
	\begin{matrix}
		r_{a_so_s}&		\cdots&		r_{a_so_e}\\
		\vdots&		\ddots&		\vdots	  \\
		r_{a_eo_s}&		\cdots&		r_{a_eo_e}\\
	\end{matrix} 
	\right] 
	\right),
\end{equation}

where $k$ is a hyperparameter, $c$ denotes the contextual information of the predicted region, which is the pooling result of the information of the corresponding region.
\subsection{Decoding and Loss function}
In decoding, a region classifier decodes potential triplets and excludes candidates with ``Invalid" sentiment labels. Potential triples are assigned two labels that record the beginning and end positions of the candidate region, as shown in Figure \ref{bdtfgrid}. The labels of the potential triplet region are $(S[a_1,o_4],E[a_3,o_5], {\rm Positive})$, where $a_i$,$o_i$ represent aspect and opinion words, respectively, and subscripts indicate the corresponding position serial numbers.

The final goal of training is to minimize the cross-entropy loss between the predicted triplet and the golden truth. Given golden truth triplet $y=(S[a_s,o_s],E[a_e,o_e],\mathcal{S})$, the predicted triplet is $\hat{y}=(S[\hat{a}_s,\hat{o}_s],E[\hat{a}_e,\hat{o}_e],\hat{\mathcal{S}})$, where $\mathcal{S} \in\{{\rm Positive,Neutral,Negative}\}$. The trained loss function $L$ can be expressed as
\begin{align}
	L_a 			&= -\underset{\hat{a}_s,\hat{a}_e\in \hat{y}}{\underset{a_s,a_e\in y}{\sum{}}}  y(a_s,a_e)\log{\hat{y}(\hat{a}_s,\hat{a}_e)}\\
	L_o 			&= -\underset{\hat{o}_s,\hat{o}_e\in \hat{y}}{\underset{o_s,o_e\in y}{\sum{}}}  y(o_s,o_e)\log{\hat{y}(\hat{o}_s,\hat{o}_e)}\\
	L_\mathcal{S}   &= -\underset{\mathcal{S},\hat{\mathcal{S}}}{\sum{}} \mathcal{S}\log{\hat{\mathcal{S}}}\\
	L 				&= L_a + L_o + L_\mathcal{S}.
\end{align}

\section{Experiments}
\subsection{Datasets}
We evaluated our framework on four benchmark datasets: L\small AP\normalsize14 and R\small EST\normalsize14 from SemEval 2014 task 4\footnote{\url{https://alt.qcri.org/semeval2014/task4/}} \cite{semeval-2014-task4}, , R\small EST\normalsize15 from SemEval 2015 task 12\footnote{\url{https://alt.qcri.org/semeval2015/task12/}} \cite{semeval-2015-task12} and , R\small EST\normalsize16 from SemEval 2016 task 5\footnote{\url{https://alt.qcri.org/semeval2016/task5/}} \cite{semeval-2016-task5}. These contain online reviews about laptop and restaurant domains. \citet{peng-two-stage} released ASTE-Data-V1 by labeling aspects and sentiments based on \citet{fan-first-label-dataset}. \citet{jet} refined these datasets and released ASTE-Data-V2. We evaluated our framework on versions V1 and V2 of the datasets, whose statistics are shown in Table \ref{dataset_statistic}, where $\#S$, $\#A$, $\#O$, and $\#T$ represent sentences, aspects, opinions, and triplets, respectively.

\subsection{Implementation}
We employed \textit{bert-base-uncased} as the underlying language model for the basic encoder. We employed Bi-LSTM and a 2-layer GCN module as components of the particular encoder. For 3-domain embedding, we employed \textit{GloVe} \citep{glove} as the domain-general embedding, and \textit{Word Embedding of Amazon Product Review Corpus} \citep{par_embedding}\footnote{\url{https://doi.org/10.5281/zenodo.3370051}} as the domain-specific embedding for comments, created using word2vec in CBOW mode, with 500 dimensions and window size 5. We randomly initialized  five 100-dimensional vectors for part-of-speech embeddings. The iteration count for the interaction module was set to 2, which we justify later in this paper. We trained the model for 15 epochs using different random seeds and selected the best-performing model on the test set based on recall (R), precision (P), and F$_1$ score.

\subsection{Comparison models}

\begin{enumerate}
	\item \textbf{Pipeline methods}
	\begin{itemize}
		\item \textbf{RINANTE+} \citep{rinante} is based on LSTM-CRF, with some mining rules added in a weakly supervised manner to extract triplets through two stages;
		\item \textbf{Peng-two-stage} \citep{peng-two-stage} is a two-stage model for extracting aspect sentiment triplets. The first stage extracts aspects, opinions, and sentiments from sentences. These pairings are combined in the second stage to produce the appropriate triplets;
		\item \textbf{BMRC} \citep{bmrc} transforms the ASTE task into a multi-turn machine reading comprehension task by associate extraction, matching, and classification by adding different queries, which include unrestricted extraction aspects and opinions, restricted aspects matching opinions, restricted opinions matching aspects, and sentiment classification.
	\end{itemize}
	
	\item \textbf{Span based methods}
	\begin{itemize}
		\item \textbf{Span-ASTE} \citep{span-aste} is based on span extraction, for an end-to-end model that cannot perform well on aspects and opinions with multiple words. After spanning the sentences, the ATE and OTE modules generate candidate aspect and opinion spans and determine possible span pairs using the matching module;
		\item \textbf{Chen-dual-decoder} \citep{chen-dual-decoder}  has two transformer-based decoders with multiple multi-head attention to improve the extraction performance of long entities and reduce cascading errors due to sequential decoding. Similarly, it is based on the span implementation.
	\end{itemize}
	
	\item \textbf{Joint extraction methods}
	\begin{itemize}
		\item \textbf{JET} \citep{jet} is a \textit{BIOES}-based position-aware tagging scheme that specifies the structures of triplets and enables the connection between aspects, opinions, and sentiment by richer tag semantics;
		\item \textbf{GTS} \citep{gts} converts a sentence into a 2D table and uses an AOSPN-based uniform grid tagging scheme to extract triplets in an end-to-end approach;
		\item \textbf{EMC-GCN} \citep{emc-gcn} uses a multi-channel graph convolutional enhanced GTS network to introduce linguistic features, including part-of-speech, syntactic dependency, tree-based distance, and relative position distance;
		\item \textbf{BDTF} \citep{bdtf} transforms ASTE into a task of detection and classification of relation regions through a boundary-driven table-filling strategy, which represents triplets as relation regions, and identifies triplets by determining the positions, starting in the upper-left and ending in the lower-right of the region.
	\end{itemize}
\end{enumerate}

\section{Results and analysis}

\subsection{Main results}
\begin{table*}
	\centering
	\scalebox{0.7}{
		\begin{tabular}{@{}lllllllllllll@{}}
			\toprule
			\multirow{2}{*}{Model} & \multicolumn{3}{c}{L\small AP\normalsize14} & \multicolumn{3}{c}{R\small EST\normalsize14} & \multicolumn{3}{c}{R\small EST\normalsize15} & \multicolumn{3}{c}{R\small EST\normalsize16} \\ \cmidrule(l){2-13} 
			& P.       & R.       & F$_1$       & P.      & R.      & F$_1$       & P.      & R.      & F$_1$       & P.      & R.      & F$_1$       \\ \midrule
			\textbf{Pipeline methods}         &          &          &          &         &         &         &         &         &         &         &         &         \\
			RINANTE+                                           & 21.71    & 18.66    & 20.07    & 31.42   & 39.38   & 34.95   & 29.88   & 30.06   & 29.97   & 25.68   & 22.30   & 23.87   \\
			Peng-two-stage                                     & 37.38    & 50.38    & 42.87    & 43.24   & 63.66   & 51.46   & 48.07   & 57.51   & 52.32   & 46.96   & 64.24   & 54.21   \\
			BMRC                                               & 65.91    & 52.15    & 58.18    & 72.17   & 65.43   & 68.64   & 62.48   & 55.55   & 58.79   & 69.87   & 65.68   & 67.35   \\ \midrule
			\textbf{Span based methods}       &          &          &          &         &         &         &         &         &         &         &         &         \\
			Span-ASTE                                          & 63.44    & 55.84    & 59.38    & 72.89   & 70.89   & 71.85   & 62.18   & 64.45   & 63.27   & 69.45   & 71.17   & 70.26   \\
			Chen-dual-decoder                                  & 62.36    & \textbf{60.37}    & 61.35    & 72.12   & 73.14   & 72.62   & 64.27   & 60.73   & 62.45   & 68.74   & 71.79   & 70.23   \\
			SSJE                                               & \underline{67.43}    & 54.71    & 60.41    & 73.12   & 71.43   & 72.26   & 63.94   & \underline{66.17}   & 65.05   & 70.82   & 72.00   & 71.38   \\ \midrule
			\textbf{Joint extraction methods} &          &          &          &         &         &         &         &         &         &         &         &         \\
			JET$^o _{m=6}$-BERT                                    & 55.39    & 47.33    & 51.04    & 70.56   & 55.94   & 62.40   & 64.45   & 51.96   & 57.53   & 70.42   & 58.37   & 63.83   \\
			GTS-BERT                                           & 57.52    & 51.92    & 54.58    & 70.92   & 69.49   & 70.20   & 59.29   & 58.07   & 58.67   & 68.58   & 66.60   & 67.58   \\
			EMC-GCN                                            & 61.70    & 56.26    & 58.81    & 71.21   & 72.39   & 71.78   & 61.54   & 62.47   & 61.93   & 65.62   & 71.30   & 68.33   \\
			BDTF                                               & \textbf{68.94}    & 55.97    & \underline{61.74}    & \underline{75.53}   & \underline{73.24}   & \underline{74.35}   & \textbf{68.76}   & 63.71   & \underline{66.12}   & \underline{71.40}   & \underline{73.13}   & \underline{72.27}   \\
			Ours   & 65.98        & \underline{58.78}        & \textbf{62.17}        & \textbf{77.14}   & \textbf{75.35}   & \textbf{76.23}   & \underline{68.07}   & \textbf{66.80}   & \textbf{67.43}   & \textbf{71.90}   & \textbf{76.65}   & \textbf{74.20}   \\ \bottomrule
		\end{tabular}
	}
	\caption{Main results on ASTE-Data-V2 dataset, the best are marked in bold and the second best results are underlined. All baseline results are from the original papers.}
	\label{main_results}
\end{table*}
\begin{table*}
	\centering
	\scalebox{0.7}{
		\begin{tabular}{@{}lllllllllllll@{}}
			\toprule
			\multirow{2}{*}{Model} & \multicolumn{3}{c}{L\small AP\normalsize14} & \multicolumn{3}{c}{R\small EST\normalsize14} & \multicolumn{3}{c}{R\small EST\normalsize15} & \multicolumn{3}{c}{R\small EST\normalsize16} \\ \cmidrule(l){2-13} 
			& P.      & R.     & F$_1$     & P.      & R.      & F$_1$     & P.      & R.      & F$_1$     & P.      & R.      & F$_1$     \\ \midrule
			\textbf{Pipeline methods}         &          &          &          &         &         &         &         &         &         &         &         &         \\
			RINANTE+                         & 23.10   & 17.60  & 20.00  & 31.07   & 37.63   & 34.03  & 29.40   & 26.90   & 28.00  & 27.10   & 20.50   & 23.30  \\
			Peng-two-stage                   & 40.40   & 47.24  & 43.50  & 44.18   & 62.99   & 51.89  & 40.97   & 54.68   & 46.79  & 46.76   & 62.97   & 53.62  \\
			BMRC                             & 65.12   & 54.41  & 59.27  & 71.32   & 70.09   & 70.69  & 63.71   & 58.63   & 61.05  & 67.74   & 68.56   & 68.13  \\ \midrule
			\textbf{Joint extraction methods}         &         &        &        &         &         &        &         &         &        &         &         &        \\
			JET$^o_{m=6}$-BERT & 58.47   & 43.67  & 50.00  & 67.97   & 60.32   & 63.92  & 58.35   & 51.43   & 54.67  & 64.77   & 61.29   & 62.98  \\
			GTS-BERT                         & 57.52   & 51.92  & 54.58  & 70.92   & 69.49   & 70.20  & 59.29   & 58.07   & 58.67  & 68.58   & 66.60   & 67.58  \\
			EMC-GCN                          & 61.46   & \underline{55.56}  & 58.32  & 71.85   & 72.12   & 71.98  & 59.89   & 61.05   & 60.38  & 65.08   & 71.66   & 68.18  \\
			BDTF                             & \textbf{68.30}   & 55.10  & \underline{60.99}  & \underline{76.71}   & \underline{74.01}   & \underline{75.33}  & \underline{66.95}   & \underline{65.05}   & \underline{65.97}  & \textbf{73.43}   & \underline{73.64}   & \underline{73.51}  \\
			Ours                             & \underline{67.05}   & \textbf{58.98}  & \textbf{62.76}  & \textbf{78.70}       & \textbf{75.02}       & \textbf{76.82}     & \textbf{68.17}   & \textbf{66.37}   & \textbf{67.26}  & \underline{70.68}       & \textbf{78.28}       & \textbf{74.29}  \\ \bottomrule
		\end{tabular}
	}
	\caption{Main results on ASTE-Data-V1 dataset, the best are marked in bold and the second best results are underlined. All baseline results are from the original papers.}
	\label{main_results_on_v1}
\end{table*}

Tables \ref{main_results} and \ref{main_results_on_v1} present the experimental results obtained from the two benchmark datasets. We evaluated the ASTE task based on precision, recall, and F$_1$-score. Based on these results, our model has a significant advantage across all datasets.

The following observations can be made.

\begin{enumerate}
	\item On ASTE-Data-V1, our model demonstrates improvements in F$_1$-score of 1.77\%, 1.49\%, 1.29\%, and 0.78\% when compared to the best joint extraction methods in $L\small AP\normalsize14, R\small EST\normalsize14, R\small EST\normalsize15$, and $R\small EST\normalsize16$, respectively.
	\item On ASTE-Data-V2, our model achieves F$_1$-score improvements of 0.43\%, 1.88\%, 1.31\%, and 1.93\% compared to the best joint extraction methods in $L\small AP\normalsize14, R\small EST\normalsize14, R\small EST\normalsize15$ and $R\small EST\normalsize16$, respectively. Our model shows improvements of 3.99\%, 7.59\%, 8.64\%, and 6.85\%, respectively, over pipeline-based methods, and 0.82\%, 3.61\%, 2.38\%, 2.82\% over span-based methods.
	\item The average improvements in precision and recall on the V1 and V2 datasets are $-$0.20, 2.60, and $-$0.39, 1.17, respectively, compared to the best values achieved by comparison models. This suggests that the enhancements in F$_1$-scores are primarily due to improvements in recall, indicating that our model excels at comprehensively extracting golden truth triplets.
\end{enumerate}

We attribute the above improvements to our dual-encoder, which incorporates a substantial amount of underlying semantic and syntactic information, enabling our model to make more accurate predictions by relying on more detailed contextual encoding.

\begin{table}
	\centering
	\scalebox{0.8}{
		\begin{tabular}{@{}lllllll@{}}
			\toprule
			\multirow{2}{*}{Method} & \multicolumn{3}{c}{L\small AP\normalsize14} & \multicolumn{3}{c}{R\small EST\normalsize14} \\ \cmidrule(l){2-7} 
			& P.      & R.     & F$_1$     & P.      & R.      & F$_1$     \\ \midrule
			Ours                    & 65.98   & 58.78  & 62.17  & 77.14   & 75.35   & 76.23  \\
			w/o Basic               & 61.45($\downarrow$4.53)   & 55.42($\downarrow$3.36)  & 58.28($\downarrow$3.89)  & 73.62($\downarrow$3.52)   & 69.74($\downarrow$5.61)   & 71.63($\downarrow$4.60)  \\
			w/o Particular          & 66.67($\uparrow$0.72)   	& 56.56($\downarrow$2.22)  & 61.20($\downarrow$0.97)  & 74.41($\downarrow$2.73)   & 72.54($\downarrow$2.81)   & 73.46($\downarrow$2.77)  \\
			w/o Interaction         & 59.23($\downarrow$6.75)   & 59.89($\uparrow$1.11)    & 59.56($\downarrow$5.61)  & 77.29($\uparrow$0.15)     & 70.52($\downarrow$4.83)   & 73.75($\downarrow$2.48)  \\
			w/o 3-domain Embedding  & 63.91($\downarrow$2.07)   & 59.89($\uparrow$1.11)    & 61.83($\downarrow$0.34)  & 74.63($\downarrow$2.51)   & 70.42($\downarrow$4.93)   & 72.46($\downarrow$3.77)  \\
			w/o GCN                 & 64.24($\downarrow$1.74)   & 58.78					   & 61.39($\downarrow$0.78)  & 77.67($\uparrow$0.53)     & 72.43($\downarrow$2.92)   & 74.96($\downarrow$1.27)  \\ \bottomrule
		\end{tabular}
	}
	\caption{Overall results of ablation experiments based on ASTE-Data-V2 L\small AP\normalsize14 and R\small EST\normalsize14. Compared to the original model, the resulting float marked with $\uparrow$ and $\downarrow$.}
	\label{ablation_study}
\end{table}

\subsection{Ablation study}

To examine the contributions of different modules, such as the encoders and the interaction module, we conducted ablation experiments on the ASTE-Data-V2 L\small AP\normalsize14 and R\small EST\normalsize14 datasets.

We evaluated the role of each encoder by removing them individually. The model without the basic encoder is denoted as w/o Basic, and that without the particular encoder as w/o Particular. Table \ref{ablation_study} presents the results of these experiments, which show substantial performance degradation when either encoder is removed. Compared to the original model's F$_1$-score, w/o Basic has reductions of 3.89 on L\small AP\normalsize14 and 4.60 on R\small EST\normalsize14, and w/o Particular has corresponding reductions of 0.97and 2.77.

We investigated the importance of the iterative interaction module in fusing high-level semantics through an ablation experiment, replacing the iterative fusion operation with tensor concatenation. The model without the encoding interaction module is denoted as w/o Interaction, which, as shown in Table \ref{ablation_study}, has reductions of the F$_1$-score of 5.61 on L\small AP\normalsize14 and 2.48 on R\small EST\normalsize14. This loss in performance can be attributed to the ineffectiveness of a simple tensor concatenation operation in integrating the two semantics, resulting in a significant loss of valuable information.

\begin{figure}
	\centering
	\scalebox{0.4}{
		\includegraphics{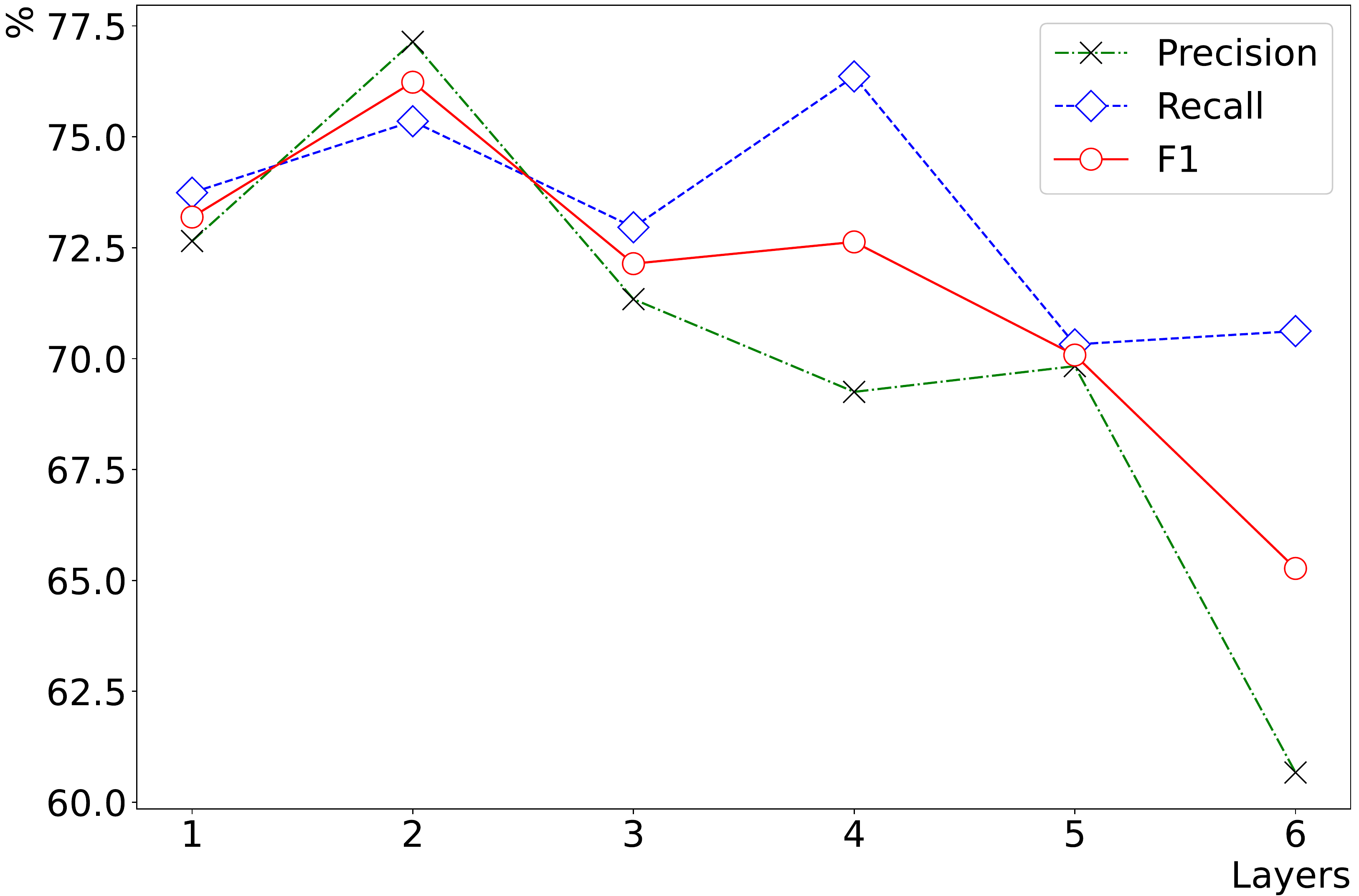}
	}
	\caption{The effect of the number of interaction layers.}
	\label{layers}
\end{figure}

\begin{table*}[]
	\scalebox{0.5}{
		\begin{tabular}{@{}lccc@{}}
			\toprule
			\multicolumn{1}{c}{Review}                                                                                              & Ground-truth                                                                                            & BDTF                                                                                                                                                & Our model                                                                                      \\ \midrule
			1. But the mountain lion is just too slow.                                                                                 & \{mountain lion, slow, NEG\}																			  & ---\XSolidBrush                                                                                                                                     & \begin{tabular}[c]{@{}c@{}}\{mountain lion, slow, NEG\}\Checkmark\\ \{mountain lion, too slow, NEG\}\XSolidBrush\end{tabular}                            \\	\midrule
			2. Strong build though which really adds to its durability.                                                                & \begin{tabular}[c]{@{}c@{}}\{build, strong, Pos\}\\ \{durability, strong, Pos\}\end{tabular}            & \{build, strong, Pos\}\Checkmark                                                                                                                    & \begin{tabular}[c]{@{}c@{}}\{build, strong, Pos\}\Checkmark\\ \{durability, strong, Pos\}\Checkmark\end{tabular} \\	\midrule
			\begin{tabular}[c]{@{}l@{}}3. The sound is nice and loud and i do n't have\\ \quad any problems with hearing anything.\end{tabular} & \begin{tabular}[c]{@{}c@{}}\{sound, nice, Pos\}\\ \{sound, loud, Pos\}\end{tabular}                     & \begin{tabular}[c]{@{}c@{}}\{sound, nice, Pos\}\Checkmark\\ \{sound, do n't have any problems, Pos\}\XSolidBrush\end{tabular}                                  & \begin{tabular}[c]{@{}c@{}}\{sound, nice, Pos\}\Checkmark\\ \{sound, loud, Pos\}\Checkmark\end{tabular}          \\ \midrule
			4. Its size is ideal and the weight is acceptable.                                                                         & \begin{tabular}[c]{@{}c@{}}\{size, ideal, Pos\}\\ \{weight, acceptable, Pos\}\end{tabular}              & \begin{tabular}[c]{@{}c@{}}\{size, acceptable, Pos\}\XSolidBrush\\ \{size, ideal, Pos\}\Checkmark\\ \{weight, acceptable, Pos\}\Checkmark\\ \{weight, ideal, Pos\}\XSolidBrush\end{tabular} & \begin{tabular}[c]{@{}c@{}}\{size, ideal, Pos\}\Checkmark\\ \{weight, acceptable, Pos\}\Checkmark\end{tabular}   \\	\midrule
			5. I really like the size and I 'm a fan of the ACERS.                                                                     & \{size, like, Pos\}                                                                                     & \begin{tabular}[c]{@{}c@{}}\{size, fan, Pos\}\XSolidBrush\\ \{size, like, Pos\}\Checkmark\\ \{ACERS, fan, Pos\}\XSolidBrush\end{tabular}                                  & \{size, like, Pos\}\Checkmark                                                                  \\	\bottomrule
		\end{tabular}
	}
	\caption{A case study based on 5 real examples from the test datasets. The correctness or incorrectness of the predicted triplets is marked by \Checkmark and \XSolidBrush.}
	\label{case_study}
\end{table*}

To explore the effect of the number of iterations in the encoding interaction module, we varied the number of iterations from 1 to 6 on the ASTE-Data-V2 R\small EST\normalsize14 dataset. We found that the optimal performance was achieved with 2 iterations, as depicted in Figure \ref{layers}. This suggests that too few or too many iterations may lead to decreased performance due to insufficient fusion or the introduction of redundant noise. Consequently, we set our model to use 2 layers of iterative interaction.

To evaluate multi-domain word embeddings in a particular encoder compared to traditional embeddings, we conducted ablation experiments using a single GloVe word embedding, referred to as w/o 3-domain, instead of the original 3-domain embedding. As shown in Table \ref{ablation_study}, its F$_1$-score is 0.34 smaller in L\small AP\normalsize14, and 3.77 smaller in R\small EST\normalsize14. This can be attributed to the inability of a single GloVe word embedding to capture the nuanced semantics present in comments, as it is limited to capturing word semantics within a universal domain. Thus, our experiments confirmed the superiority of multi-domain embeddings for semantic extraction.

Finally, to investigate the impact of text dependencies on the results, we conducted an ablation experiment by removing the dependency tree-based graph convolution module, denoted as w/o GCN. As shown in Table \ref{ablation_study}, the F$_1$-score of the model decreases by 0.78 in L\small AP\normalsize14 and 1.27 in R\small EST\normalsize14, which can be attributed to the absence of the GCN module, which hinders the model's ability to capture grammar-based structural information, and subsequently impairs its predictive accuracy. This emphasizes the crucial role of grammatical structure information in semantic extraction.

\subsection{Case analysis}

We compared our model with the state-of-the-art model BDTF using five real-world cases from the datasets. The results are presented in Table \ref{case_study}, highlighting our model's superior performance in the majority of cases.

In sentence 1, our model, which was pretrained on the laptop dataset, accurately predicts the sentiment triplet for a comment that is not explicitly about laptops, while BDTF fails to capture the sentiment triplet in this case. This showcases the strong robustness of our model, enabling it to adapt to more complex testing scenarios.

For sentences 2 and 3, where multiple sentiment triplets are present in a single sentence, our model demonstrates a more comprehensive retrieval of results compared to BDTF. While both models achieve correct results, our model shows greater precision.

In sentences 4 and 5, both models achieve correct predictions. However, BDTF produces a significant number of erroneous predictions, while ours maintains a higher precision rate. This highlights our model’s ability to mitigate error propagation and deliver more robust predictions.

\section{Conclusion}
We proposed a semantically enhanced dual-encoder for aspect sentiment triplet extraction. Unlike previous work, we encoded more detailed semantics, which enabled improvements and an excellent table-filling strategy. To ensure accurate and comprehensive encoding, we designed a BERT-based basic encoder and GCN-BiLSTM-based particular encoder. We introduced an iterative interaction module to fuse the two semantic encodings, resulting in a more appropriate semantic representation. Experiments and ablation studies demonstrated the superiority and robustness of our model. Next, we will focus on extracting higher-order semantic information to further improve the performance of the baseline model.

\section{CRediT authorship contribution statement}
\textbf{Baoxing Jiang:} Conceptualization, Methodology, Software, Validation, Writing - Original Draft.  \textbf{Shehui Liang:} Funding acquisition, Project administration. \textbf{Peiyu Liu:} Project administration, Supervision. \textbf{Kaifang Dong:} Writing - Review \& Editing, Formal analysis. \textbf{Hongye Li:} Writing - Review \& Editing, Data Curation.

\section{Declaration of competing interest}
The authors declare that they have no known competing financial interests or personal relationships that could have appeared
to influence the work reported in this paper.

\section{Acknowledgments}
This work was supported by Jiangsu Provincial Social Science Foundation of China (21HQ043), Key R \& D project of Shandong Province (2019JZZY010129), Shandong Provincial Social Science Planning Project under Award (19BJCJ51,18CXWJ01, 18BJYJ04)

\bibliographystyle{elsarticle-num-names} 
\bibliography{reference}

\begin{thebibliography}{36}
\expandafter\ifx\csname natexlab\endcsname\relax\def\natexlab#1{#1}\fi
\providecommand{\url}[1]{\texttt{#1}}
\providecommand{\href}[2]{#2}
\providecommand{\path}[1]{#1}
\providecommand{\DOIprefix}{doi:}
\providecommand{\ArXivprefix}{arXiv:}
\providecommand{\URLprefix}{URL: }
\providecommand{\Pubmedprefix}{pmid:}
\providecommand{\doi}[1]{\href{http://dx.doi.org/#1}{\path{#1}}}
\providecommand{\Pubmed}[1]{\href{pmid:#1}{\path{#1}}}
\providecommand{\bibinfo}[2]{#2}
\ifx\xfnm\relax \def\xfnm[#1]{\unskip,\space#1}\fi
\bibitem[{Peng et~al.(2020)Peng, Xu, Bing, Huang, Lu, and Si}]{peng-two-stage}
\bibinfo{author}{H.~Peng}, \bibinfo{author}{L.~Xu}, \bibinfo{author}{L.~Bing},
  \bibinfo{author}{F.~Huang}, \bibinfo{author}{W.~Lu}, \bibinfo{author}{L.~Si},
\newblock \bibinfo{title}{Knowing what, how and why: A near complete solution
  for aspect-based sentiment analysis},
\newblock in: \bibinfo{booktitle}{The Thirty-Fourth {AAAI} Conference on
  Artificial Intelligence (AAAI)}, \bibinfo{publisher}{{AAAI} Press},
  \bibinfo{address}{New York, NY, USA}, \bibinfo{year}{2020}, pp.
  \bibinfo{pages}{8600--8607}. \DOIprefix\doi{10.1609/aaai.v34i05.6383}.
\bibitem[{Pontiki et~al.(2014)Pontiki, Galanis, Pavlopoulos, Papageorgiou,
  Androutsopoulos, and Manandhar}]{semeval-2014-task4}
\bibinfo{author}{M.~Pontiki}, \bibinfo{author}{D.~Galanis},
  \bibinfo{author}{J.~Pavlopoulos}, \bibinfo{author}{H.~Papageorgiou},
  \bibinfo{author}{I.~Androutsopoulos}, \bibinfo{author}{S.~Manandhar},
\newblock \bibinfo{title}{Semeval-2014 task 4: Aspect based sentiment
  analysis},
\newblock in: \bibinfo{booktitle}{Proceedings of the 8th International Workshop
  on Semantic Evaluation (SemEval{@}COLING)}, \bibinfo{publisher}{ACL},
  \bibinfo{address}{Dublin, Ireland}, \bibinfo{year}{2014}, pp.
  \bibinfo{pages}{27--35}. \DOIprefix\doi{10.3115/v1/s14-2004}.
\bibitem[{He et~al.(2019)He, Lee, Ng, and Dahlmeier}]{ate1}
\bibinfo{author}{R.~He}, \bibinfo{author}{W.~S. Lee}, \bibinfo{author}{H.~T.
  Ng}, \bibinfo{author}{D.~Dahlmeier},
\newblock \bibinfo{title}{An interactive multi-task learning network for
  end-to-end aspect-based sentiment analysis},
\newblock in: \bibinfo{booktitle}{Proceedings of the 57th Conference of the
  Association for Computational Linguistics (ACL)}, \bibinfo{publisher}{ACL},
  \bibinfo{address}{Florence, Italy}, \bibinfo{year}{2019}, pp.
  \bibinfo{pages}{504--515}. \DOIprefix\doi{10.18653/v1/p19-1048}.
\bibitem[{Wu et~al.(2020)Wu, Zhao, Dai, Huang, and Chen}]{ate2}
\bibinfo{author}{Z.~Wu}, \bibinfo{author}{F.~Zhao}, \bibinfo{author}{X.~Dai},
  \bibinfo{author}{S.~Huang}, \bibinfo{author}{J.~Chen},
\newblock \bibinfo{title}{Latent opinions transfer network for target-oriented
  opinion words extraction},
\newblock in: \bibinfo{booktitle}{The Thirty-Fourth {AAAI} Conference on
  Artificial Intelligence (AAAI), The Thirty-Second Innovative Applications of
  Artificial Intelligence Conference (IAAI), The Tenth {AAAI} Symposium on
  Educational Advances in Artificial Intelligence (EAAI)},
  \bibinfo{publisher}{AAAI Press}, \bibinfo{address}{New York, NY, USA},
  \bibinfo{year}{2020}, pp. \bibinfo{pages}{9298--9305}.
  \DOIprefix\doi{10.1609/aaai.v34i05.6469}.
\bibitem[{Ma et~al.(2019)Ma, Li, Wu, Xie, and Wang}]{ate3}
\bibinfo{author}{D.~Ma}, \bibinfo{author}{S.~Li}, \bibinfo{author}{F.~Wu},
  \bibinfo{author}{X.~Xie}, \bibinfo{author}{H.~Wang},
\newblock \bibinfo{title}{Exploring sequence-to-sequence learning in aspect
  term extraction},
\newblock in: \bibinfo{booktitle}{Proceedings of the 57th Conference of the
  Association for Computational Linguistics (ACL)}, \bibinfo{publisher}{ACL},
  \bibinfo{address}{Florence, Italy}, \bibinfo{year}{2019}, pp.
  \bibinfo{pages}{3538--3547}. \DOIprefix\doi{10.18653/v1/p19-1344}.
\bibitem[{Zhou et~al.(2020)Zhou, Jiang, Song, Su, Guo, Han, and Hu}]{ote1}
\bibinfo{author}{Y.~Zhou}, \bibinfo{author}{W.~Jiang},
  \bibinfo{author}{P.~Song}, \bibinfo{author}{Y.~Su}, \bibinfo{author}{T.~Guo},
  \bibinfo{author}{J.~Han}, \bibinfo{author}{S.~Hu},
\newblock \bibinfo{title}{Graph convolutional networks for target-oriented
  opinion words extraction with adversarial training},
\newblock in: \bibinfo{booktitle}{2020 International Joint Conference on Neural
  Networks (IJCNN)}, \bibinfo{publisher}{IEEE}, \bibinfo{address}{Glasgow,
  United Kingdom}, \bibinfo{year}{2020}, pp. \bibinfo{pages}{1--7}.
  \DOIprefix\doi{10.1109/IJCNN48605.2020.9207463}.
\bibitem[{Xing et~al.(2023)Xing, Zhu, Fan, Zhang, Huang, Gu, Ip, and
  Yung}]{ote2}
\bibinfo{author}{Y.~Xing}, \bibinfo{author}{Y.~Zhu}, \bibinfo{author}{W.~Fan},
  \bibinfo{author}{Y.~Zhang}, \bibinfo{author}{R.~Huang},
  \bibinfo{author}{Z.~Gu}, \bibinfo{author}{W.~H. Ip},
  \bibinfo{author}{K.~Yung},
\newblock \bibinfo{title}{Spanmtl: a span-based multi-table labeling for
  aspect-oriented fine-grained opinion extraction},
\newblock \bibinfo{journal}{Soft Computing} \bibinfo{volume}{27}
  (\bibinfo{year}{2023}) \bibinfo{pages}{4627--4637}.
  \DOIprefix\doi{10.1007/s00500-022-07721-5}.
\bibitem[{Kang et~al.(2022)Kang, Kim, Yun, Lee, and Jung}]{aoe1}
\bibinfo{author}{T.~Kang}, \bibinfo{author}{S.~Kim}, \bibinfo{author}{H.~Yun},
  \bibinfo{author}{H.~Lee}, \bibinfo{author}{K.~Jung},
\newblock \bibinfo{title}{Gated relational encoder-decoder model for
  target-oriented opinion word extraction},
\newblock \bibinfo{journal}{{IEEE} Access} \bibinfo{volume}{10}
  (\bibinfo{year}{2022}) \bibinfo{pages}{130507--130517}.
  \DOIprefix\doi{10.1109/ACCESS.2022.3228835}.
\bibitem[{Liu et~al.(2022)Liu, Li, Fei, and Ji}]{aoe2}
\bibinfo{author}{Y.~Liu}, \bibinfo{author}{F.~Li}, \bibinfo{author}{H.~Fei},
  \bibinfo{author}{D.~Ji},
\newblock \bibinfo{title}{Pair-wise aspect and opinion terms extraction as
  graph parsing via a novel mutually-aware interaction mechanism},
\newblock \bibinfo{journal}{Neurocomputing} \bibinfo{volume}{493}
  (\bibinfo{year}{2022}) \bibinfo{pages}{268--280}.
  \DOIprefix\doi{10.1016/j.neucom.2022.04.064}.
\bibitem[{Zhang et~al.(2022)Zhang, Peng, Han, Han, Yue, and Liu}]{aoe3}
\bibinfo{author}{Y.~Zhang}, \bibinfo{author}{T.~Peng},
  \bibinfo{author}{R.~Han}, \bibinfo{author}{J.~Han}, \bibinfo{author}{L.~Yue},
  \bibinfo{author}{L.~Liu},
\newblock \bibinfo{title}{Synchronously tracking entities and relations in a
  syntax-aware parallel architecture for aspect-opinion pair extraction},
\newblock \bibinfo{journal}{Applied Intelligence} \bibinfo{volume}{52}
  (\bibinfo{year}{2022}) \bibinfo{pages}{15210--15225}.
  \DOIprefix\doi{10.1007/s10489-022-03286-w}.
\bibitem[{Jiang et~al.(2023)Jiang, Xu, and Liu}]{alsc1}
\bibinfo{author}{B.~Jiang}, \bibinfo{author}{G.~Xu}, \bibinfo{author}{P.~Liu},
\newblock \bibinfo{title}{Aspect-level sentiment classification via location
  enhanced aspect-merged graph convolutional networks},
\newblock \bibinfo{journal}{J. Supercomput.} \bibinfo{volume}{79}
  (\bibinfo{year}{2023}) \bibinfo{pages}{9666--9691}.
  \DOIprefix\doi{10.1007/s11227-022-05002-4}.
\bibitem[{Huang et~al.(2023)Huang, Peng, Liu, Yang, Wang,
  Orellana{-}Mart{\'{\i}}n, and P{\'{e}}rez{-}Jim{\'{e}}nez}]{alsc2}
\bibinfo{author}{Y.~Huang}, \bibinfo{author}{H.~Peng},
  \bibinfo{author}{Q.~Liu}, \bibinfo{author}{Q.~Yang},
  \bibinfo{author}{J.~Wang}, \bibinfo{author}{D.~Orellana{-}Mart{\'{\i}}n},
  \bibinfo{author}{M.~J. P{\'{e}}rez{-}Jim{\'{e}}nez},
\newblock \bibinfo{title}{Attention-enabled gated spiking neural p model for
  aspect-level sentiment classification},
\newblock \bibinfo{journal}{Neural Networks} \bibinfo{volume}{157}
  (\bibinfo{year}{2023}) \bibinfo{pages}{437--443}.
  \DOIprefix\doi{10.1016/j.neunet.2022.11.006}.
\bibitem[{Zhang et~al.(2023)Zhang, Xu, Cai, Tan, and Zhu}]{alsc3}
\bibinfo{author}{X.~Zhang}, \bibinfo{author}{J.~Xu}, \bibinfo{author}{Y.~Cai},
  \bibinfo{author}{X.~Tan}, \bibinfo{author}{C.~Zhu},
\newblock \bibinfo{title}{Detecting dependency-related sentiment features for
  aspect-level sentiment classification},
\newblock \bibinfo{journal}{IEEE Transactions on Affective Computing}
  \bibinfo{volume}{14} (\bibinfo{year}{2023}) \bibinfo{pages}{196--210}.
  \DOIprefix\doi{10.1109/TAFFC.2021.3063259}.
\bibitem[{Xu et~al.(2020)Xu, Li, Lu, and Bing}]{jet}
\bibinfo{author}{L.~Xu}, \bibinfo{author}{H.~Li}, \bibinfo{author}{W.~Lu},
  \bibinfo{author}{L.~Bing},
\newblock \bibinfo{title}{Position-aware tagging for aspect sentiment triplet
  extraction},
\newblock in: \bibinfo{booktitle}{Proceedings of the 2020 Conference on
  Empirical Methods in Natural Language Processing (EMNLP)},
  \bibinfo{publisher}{ACL}, \bibinfo{address}{Onlien}, \bibinfo{year}{2020},
  pp. \bibinfo{pages}{2339--2349}.
  \DOIprefix\doi{10.18653/v1/2020.emnlp-main.183}.
\bibitem[{Xu et~al.(2021)Xu, Chia, and Bing}]{span-aste}
\bibinfo{author}{L.~Xu}, \bibinfo{author}{Y.~K. Chia},
  \bibinfo{author}{L.~Bing},
\newblock \bibinfo{title}{Learning span-level interactions for aspect sentiment
  triplet extraction},
\newblock in: \bibinfo{booktitle}{Proceedings of the 59th Annual Meeting of the
  Association for Computational Linguistics (ACL), The 11th International Joint
  Conference on Natural Language Processing (IJCNLP)},
  \bibinfo{publisher}{ACL}, \bibinfo{address}{Online}, \bibinfo{year}{2021},
  pp. \bibinfo{pages}{4755--4766}.
  \DOIprefix\doi{10.18653/v1/2021.acl-long.367}.
\bibitem[{Wu et~al.(2020)Wu, Ying, Zhao, Fan, Dai, and Xia}]{gts}
\bibinfo{author}{Z.~Wu}, \bibinfo{author}{C.~Ying}, \bibinfo{author}{F.~Zhao},
  \bibinfo{author}{Z.~Fan}, \bibinfo{author}{X.~Dai}, \bibinfo{author}{R.~Xia},
\newblock \bibinfo{title}{Grid tagging scheme for aspect-oriented fine-grained
  opinion extraction},
\newblock in: \bibinfo{booktitle}{Findings of the Association for Computational
  Linguistics (EMNLP)}, \bibinfo{publisher}{ACL}, \bibinfo{address}{Online},
  \bibinfo{year}{2020}, pp. \bibinfo{pages}{2576--2585}.
  \DOIprefix\doi{10.18653/v1/2020.findings-emnlp.234}.
\bibitem[{Chen et~al.(2022)Chen, Zhai, Feng, Li, and Wang}]{emc-gcn}
\bibinfo{author}{H.~Chen}, \bibinfo{author}{Z.~Zhai},
  \bibinfo{author}{F.~Feng}, \bibinfo{author}{R.~Li},
  \bibinfo{author}{X.~Wang},
\newblock \bibinfo{title}{Enhanced multi-channel graph convolutional network
  for aspect sentiment triplet extraction},
\newblock in: \bibinfo{booktitle}{Proceedings of the 60th Annual Meeting of the
  Association for Computational Linguistics (ACL)}, \bibinfo{publisher}{ACL},
  \bibinfo{address}{Dublin, Ireland}, \bibinfo{year}{2022}, pp.
  \bibinfo{pages}{2974--2985}. \DOIprefix\doi{10.18653/v1/2022.acl-long.212}.
\bibitem[{Mao et~al.(2021)Mao, Shen, Yu, and Cai}]{mao-joint}
\bibinfo{author}{Y.~Mao}, \bibinfo{author}{Y.~Shen}, \bibinfo{author}{C.~Yu},
  \bibinfo{author}{L.~Cai},
\newblock \bibinfo{title}{A joint training dual-mrc framework for aspect based
  sentiment analysis},
\newblock in: \bibinfo{booktitle}{Thirty-Fifth {AAAI} Conference on Artificial
  Intelligence (AAAI)}, \bibinfo{publisher}{AAAI Press},
  \bibinfo{address}{Online}, \bibinfo{year}{2021}, pp.
  \bibinfo{pages}{13543--13551}. \DOIprefix\doi{10.1609/aaai.v35i15.17597}.
\bibitem[{Jing et~al.(2021)Jing, Li, Zhao, and Jiang}]{jing-joint}
\bibinfo{author}{H.~Jing}, \bibinfo{author}{Z.~Li}, \bibinfo{author}{H.~Zhao},
  \bibinfo{author}{S.~Jiang},
\newblock \bibinfo{title}{Seeking common but distinguishing difference, {A}
  joint aspect-based sentiment analysis model},
\newblock in: \bibinfo{booktitle}{Proceedings of the 2021 Conference on
  Empirical Methods in Natural Language Processing (EMNLP)},
  \bibinfo{publisher}{ACL}, \bibinfo{address}{Punta Cana, Dominican Republic},
  \bibinfo{year}{2021}, pp. \bibinfo{pages}{3910--3922}.
  \DOIprefix\doi{10.18653/v1/2021.emnlp-main.318}.
\bibitem[{Zhang et~al.(2022)Zhang, Yang, Li, Liang, Chen, Dang, Yang, and
  Xu}]{bdtf}
\bibinfo{author}{Y.~Zhang}, \bibinfo{author}{Y.~Yang}, \bibinfo{author}{Y.~Li},
  \bibinfo{author}{B.~Liang}, \bibinfo{author}{S.~Chen},
  \bibinfo{author}{Y.~Dang}, \bibinfo{author}{M.~Yang},
  \bibinfo{author}{R.~Xu},
\newblock \bibinfo{title}{Boundary-driven table-filling for aspect sentiment
  triplet extraction},
\newblock in: \bibinfo{booktitle}{Proceedings of the 2022 Conference on
  Empirical Methods in Natural Language Processing (EMNLP)},
  \bibinfo{publisher}{ACL}, \bibinfo{address}{Abu Dhabi, United Arab Emirates},
  \bibinfo{year}{2022}, pp. \bibinfo{pages}{6485--6498}. \URLprefix
  \url{https://aclanthology.org/2022.emnlp-main.435}.
\bibitem[{Shi et~al.(2022)Shi, Han, Han, Qiao, and Wu}]{dependencyenhanced}
\bibinfo{author}{L.~Shi}, \bibinfo{author}{D.~Han}, \bibinfo{author}{J.~Han},
  \bibinfo{author}{B.~Qiao}, \bibinfo{author}{G.~Wu},
\newblock \bibinfo{title}{Dependency graph enhanced interactive attention
  network for aspect sentiment triplet extraction},
\newblock \bibinfo{journal}{Neurocomputing} \bibinfo{volume}{507}
  (\bibinfo{year}{2022}) \bibinfo{pages}{315--324}.
  \DOIprefix\doi{10.1016/j.neucom.2022.07.067}.
\bibitem[{Shuang et~al.(2021)Shuang, Gu, Li, Loo, and Su}]{posaware}
\bibinfo{author}{K.~Shuang}, \bibinfo{author}{M.~Gu}, \bibinfo{author}{R.~Li},
  \bibinfo{author}{J.~Loo}, \bibinfo{author}{S.~Su},
\newblock \bibinfo{title}{Interactive pos-aware network for aspect-level
  sentiment classification},
\newblock \bibinfo{journal}{Neurocomputing} \bibinfo{volume}{420}
  (\bibinfo{year}{2021}) \bibinfo{pages}{181--196}.
  \DOIprefix\doi{10.1016/j.neucom.2020.08.013}.
\bibitem[{Hochreiter and Schmidhuber(1997)}]{lstm}
\bibinfo{author}{S.~Hochreiter}, \bibinfo{author}{J.~Schmidhuber},
\newblock \bibinfo{title}{Long short-term memory},
\newblock \bibinfo{journal}{Neural Computation} \bibinfo{volume}{9}
  (\bibinfo{year}{1997}) \bibinfo{pages}{1735--1780}.
  \DOIprefix\doi{10.1162/neco.1997.9.8.1735}.
\bibitem[{Devlin et~al.(2019)Devlin, Chang, Lee, and Toutanova}]{bert}
\bibinfo{author}{J.~Devlin}, \bibinfo{author}{M.~Chang},
  \bibinfo{author}{K.~Lee}, \bibinfo{author}{K.~Toutanova},
\newblock \bibinfo{title}{{BERT:} pre-training of deep bidirectional
  transformers for language understanding},
\newblock in: \bibinfo{booktitle}{Proceedings of the 2019 Conference of the
  North American Chapter of the Association for Computational Linguistics
  (NAACL)}, \bibinfo{publisher}{ACL}, \bibinfo{address}{Minneapolis, MN, USA},
  \bibinfo{year}{2019}, pp. \bibinfo{pages}{4171--4186}.
  \DOIprefix\doi{10.18653/v1/n19-1423}.
\bibitem[{Kipf and Welling(2017)}]{gcn}
\bibinfo{author}{T.~N. Kipf}, \bibinfo{author}{M.~Welling},
\newblock \bibinfo{title}{Semi-supervised classification with graph
  convolutional networks},
\newblock in: \bibinfo{booktitle}{5th International Conference on Learning
  Representations (ICLR)}, \bibinfo{publisher}{OpenReview.net},
  \bibinfo{address}{Toulon, France}, \bibinfo{year}{2017}, pp.
  \bibinfo{pages}{1--14}. \DOIprefix\doi{10.48550/arXiv.1609.02907}.
\bibitem[{Dai et~al.(2022)Dai, Chen, Xia, Wang, and Chen}]{doubleembedding}
\bibinfo{author}{D.~Dai}, \bibinfo{author}{T.~Chen}, \bibinfo{author}{S.~Xia},
  \bibinfo{author}{G.~Wang}, \bibinfo{author}{Z.~Chen},
\newblock \bibinfo{title}{Double embedding and bidirectional sentiment
  dependence detector for aspect sentiment triplet extraction},
\newblock \bibinfo{journal}{Knowledge-Based Systems} \bibinfo{volume}{253}
  (\bibinfo{year}{2022}) \bibinfo{pages}{109506}.
  \DOIprefix\doi{10.1016/j.knosys.2022.109506}.
\bibitem[{Hendrycks and Gimpel(2016)}]{gelu}
\bibinfo{author}{D.~Hendrycks}, \bibinfo{author}{K.~Gimpel},
\newblock \bibinfo{title}{Gaussian error linear units (gelus)},
\newblock \bibinfo{journal}{arXiv e-prints}  (\bibinfo{year}{2016})
  \bibinfo{pages}{arXiv--1606}. \DOIprefix\doi{10.48550/arXiv.1606.08415}.
\bibitem[{He et~al.(2016)He, Zhang, Ren, and Sun}]{resnet}
\bibinfo{author}{K.~He}, \bibinfo{author}{X.~Zhang}, \bibinfo{author}{S.~Ren},
  \bibinfo{author}{J.~Sun},
\newblock \bibinfo{title}{Deep residual learning for image recognition},
\newblock in: \bibinfo{booktitle}{2016 {IEEE} Conference on Computer Vision and
  Pattern Recognition (CVPR)}, \bibinfo{publisher}{{IEEE} Computer Society},
  \bibinfo{address}{Las Vegas, NV, USA}, \bibinfo{year}{2016}, pp.
  \bibinfo{pages}{770--778}. \DOIprefix\doi{10.1109/CVPR.2016.90}.
\bibitem[{Pontiki et~al.(2015)Pontiki, Galanis, Papageorgiou, Manandhar, and
  Androutsopoulos}]{semeval-2015-task12}
\bibinfo{author}{M.~Pontiki}, \bibinfo{author}{D.~Galanis},
  \bibinfo{author}{H.~Papageorgiou}, \bibinfo{author}{S.~Manandhar},
  \bibinfo{author}{I.~Androutsopoulos},
\newblock \bibinfo{title}{Semeval-2015 task 12: Aspect based sentiment
  analysis},
\newblock in: \bibinfo{booktitle}{Proceedings of the 9th International Workshop
  on Semantic Evaluation (SemEval{@}NAACL-HLT)}, \bibinfo{publisher}{ACL},
  \bibinfo{address}{Denver, Colorado, USA}, \bibinfo{year}{2015}, pp.
  \bibinfo{pages}{486--495}. \DOIprefix\doi{10.18653/v1/s15-2082}.
\bibitem[{Pontiki et~al.(2016)Pontiki, Galanis, Papageorgiou, Androutsopoulos,
  Manandhar, Al{-}Smadi, Al{-}Ayyoub, Zhao, Qin, Clercq, Hoste, Apidianaki,
  Tannier, Loukachevitch, Kotelnikov, Bel, Zafra, and
  Eryigit}]{semeval-2016-task5}
\bibinfo{author}{M.~Pontiki}, \bibinfo{author}{D.~Galanis},
  \bibinfo{author}{H.~Papageorgiou}, \bibinfo{author}{I.~Androutsopoulos},
  \bibinfo{author}{S.~Manandhar}, \bibinfo{author}{M.~Al{-}Smadi},
  \bibinfo{author}{M.~Al{-}Ayyoub}, \bibinfo{author}{Y.~Zhao},
  \bibinfo{author}{B.~Qin}, \bibinfo{author}{O.~D. Clercq},
  \bibinfo{author}{V.~Hoste}, \bibinfo{author}{M.~Apidianaki},
  \bibinfo{author}{X.~Tannier}, \bibinfo{author}{N.~V. Loukachevitch},
  \bibinfo{author}{E.~V. Kotelnikov}, \bibinfo{author}{N.~Bel},
  \bibinfo{author}{S.~M.~J. Zafra}, \bibinfo{author}{G.~Eryigit},
\newblock \bibinfo{title}{Semeval-2016 task 5: Aspect based sentiment
  analysis},
\newblock in: \bibinfo{booktitle}{Proceedings of the 10th International
  Workshop on Semantic Evaluation (SemEval{@}NAACL-HLT)},
  \bibinfo{publisher}{ACL}, \bibinfo{address}{San Diego, CA, USA},
  \bibinfo{year}{2016}, pp. \bibinfo{pages}{19--30}.
  \DOIprefix\doi{10.18653/v1/s16-1002}.
\bibitem[{Fan et~al.(2019)Fan, Wu, Dai, Huang, and
  Chen}]{fan-first-label-dataset}
\bibinfo{author}{Z.~Fan}, \bibinfo{author}{Z.~Wu}, \bibinfo{author}{X.~Dai},
  \bibinfo{author}{S.~Huang}, \bibinfo{author}{J.~Chen},
\newblock \bibinfo{title}{Target-oriented opinion words extraction with
  target-fused neural sequence labeling},
\newblock in: \bibinfo{booktitle}{Proceedings of the 2019 Conference of the
  North American Chapter of the Association for Computational Linguistics
  (NAACL)}, \bibinfo{publisher}{ACL}, \bibinfo{address}{Minneapolis, MN, USA},
  \bibinfo{year}{2019}, pp. \bibinfo{pages}{2509--2518}.
  \DOIprefix\doi{10.18653/v1/n19-1259}.
\bibitem[{Pennington et~al.(2014)Pennington, Socher, and Manning}]{glove}
\bibinfo{author}{J.~Pennington}, \bibinfo{author}{R.~Socher},
  \bibinfo{author}{C.~D. Manning},
\newblock \bibinfo{title}{Glove: Global vectors for word representation},
\newblock in: \bibinfo{booktitle}{Proceedings of the 2014 Conference on
  Empirical Methods in Natural Language Processing (EMNLP)},
  \bibinfo{publisher}{ACL}, \bibinfo{address}{Doha, Qatar},
  \bibinfo{year}{2014}, pp. \bibinfo{pages}{1532--1543}.
  \DOIprefix\doi{10.3115/v1/d14-1162}.
\bibitem[{Schulder et~al.(2017)Schulder, Wiegand, Ruppenhofer, and
  Roth}]{par_embedding}
\bibinfo{author}{M.~Schulder}, \bibinfo{author}{M.~Wiegand},
  \bibinfo{author}{J.~Ruppenhofer}, \bibinfo{author}{B.~Roth},
\newblock \bibinfo{title}{Towards bootstrapping a polarity shifter lexicon
  using linguistic features},
\newblock in: \bibinfo{booktitle}{Proceedings of the Eighth International Joint
  Conference on Natural Language Processing (IJCNLP)},
  \bibinfo{publisher}{Asian Federation of Natural Language Processing},
  \bibinfo{address}{Taipei, Taiwan}, \bibinfo{year}{2017}, pp.
  \bibinfo{pages}{624--633}. \URLprefix
  \url{https://aclanthology.org/I17-1063/}.
\bibitem[{Dai and Song(2019)}]{rinante}
\bibinfo{author}{H.~Dai}, \bibinfo{author}{Y.~Song},
\newblock \bibinfo{title}{Neural aspect and opinion term extraction with mined
  rules as weak supervision},
\newblock in: \bibinfo{booktitle}{Proceedings of the 57th Annual Meeting of the
  Association for Computational Linguistics (ACL)}, \bibinfo{publisher}{ACL},
  \bibinfo{address}{Florence, Italy}, \bibinfo{year}{2019}, pp.
  \bibinfo{pages}{5268--5277}. \DOIprefix\doi{10.18653/v1/p19-1520}.
\bibitem[{Chen et~al.(2021)Chen, Wang, Liu, and Wang}]{bmrc}
\bibinfo{author}{S.~Chen}, \bibinfo{author}{Y.~Wang}, \bibinfo{author}{J.~Liu},
  \bibinfo{author}{Y.~Wang},
\newblock \bibinfo{title}{Bidirectional machine reading comprehension for
  aspect sentiment triplet extraction},
\newblock in: \bibinfo{booktitle}{Thirty-Fifth {AAAI} Conference on Artificial
  Intelligence (AAAI), Thirty-Third Conference on Innovative Applications of
  Artificial Intelligence (IAAI), The Eleventh Symposium on Educational
  Advances in Artificial Intelligence (EAAI)}, \bibinfo{publisher}{{AAAI}
  Press}, \bibinfo{address}{Online}, \bibinfo{year}{2021}, pp.
  \bibinfo{pages}{12666--12674}. \DOIprefix\doi{10.1609/aaai.v35i14.17500}.
\bibitem[{Chen et~al.(2022)Chen, Zhang, Zhou, Sun, and
  Chen}]{chen-dual-decoder}
\bibinfo{author}{Y.~Chen}, \bibinfo{author}{Z.~Zhang},
  \bibinfo{author}{G.~Zhou}, \bibinfo{author}{X.~Sun},
  \bibinfo{author}{K.~Chen},
\newblock \bibinfo{title}{Span-based dual-decoder framework for aspect
  sentiment triplet extraction},
\newblock \bibinfo{journal}{Neurocomputing} \bibinfo{volume}{492}
  (\bibinfo{year}{2022}) \bibinfo{pages}{211--221}.
  \DOIprefix\doi{10.1016/j.neucom.2022.04.022}.

\end{thebibliography}

\end{document}